\documentclass[sigconf]{acmart}
\usepackage{multirow}
\usepackage{enumitem}
\usepackage{algorithm}
\usepackage{algpseudocode}

\newcommand{\smallsection}[1]{{\vspace{0.05in} \noindent \bf {#1.\hspace{5pt}}}}
\AtBeginDocument{%
  }

\copyrightyear{2026}
\acmYear{2026}
\setcopyright{cc}
\setcctype{by}
\acmConference[WWW '26]{Proceedings of the ACM Web Conference 2026}{April 13--17, 2026}{Dubai, United Arab Emirates}
\acmBooktitle{Proceedings of the ACM Web Conference 2026 (WWW '26), April 13--17, 2026, Dubai, United Arab Emirates}
\acmPrice{}
\acmDOI{10.1145/3774904.3792731}
\acmISBN{979-8-4007-2307-0/2026/04}
\begin{document}

\title{Dynamic Multi-period Experts for Online Time Series Forecasting}

\author{Seungha Hong}
\email{shhong97@postech.ac.kr}
\affiliation{%
  \institution{Pohang University
of Science and Technology}
  \city{Pohang}
  \country{Republic of Korea}
}

\author{Sukang Chae}
\email{chaesgng2@postech.ac.kr}
\affiliation{%
  \institution{Pohang University
of Science and Technology}
  \city{Pohang}
  \country{Republic of Korea}
}

\author{Suyeon Kim}
\email{kimsu55@knu.ac.kr}
\affiliation{%
  \institution{Kyungpook National University}
  \city{Daegu}
  \country{Republic of Korea}
}

\author{Sanghwan Jang}
\email{s.jang@postech.ac.kr}
\affiliation{%
  \institution{Pohang University
of Science and Technology}
  \city{Pohang}
  \country{Republic of Korea}
}

\author{Hwanjo Yu}
\authornote{Corresponding author.}
\email{hwanjoyu@postech.ac.kr}
\affiliation{
  \institution{Pohang University
of Science and Technology}
  \city{Pohang}
  \country{Republic of Korea}
}


\begin{abstract}
Online Time Series Forecasting (OTSF) requires models to continuously adapt to concept drift. However, existing methods often treat concept drift as a monolithic phenomenon. To address this limitation, we first redefine concept drift by categorizing it into two distinct types: Recurring Drift, where previously seen patterns reappear, and Emergent Drift, where entirely new patterns emerge. We then propose DynaME (Dynamic Multi-period Experts), a novel hybrid framework designed to effectively address this dual nature of drift. For Recurring Drift, DynaME employs a committee of specialized experts that are dynamically fitted to the most relevant historical periodic patterns at each time step. For Emergent Drift, the framework detects high-uncertainty scenarios and shifts reliance to a stable, general expert. Extensive experiments on several benchmark datasets and backbones demonstrate that DynaME effectively adapts to both concept drifts and significantly outperforms existing baselines. Our source code is available at \url{https://github.com/shhong97/DynaME}.

\end{abstract}




\maketitle

\section{Introduction}

Time series forecasting (TSF) is a critical task across numerous domains, such as efficient resource allocation in power grids, weather forecasting, and informing financial trading strategies \cite{survey}. The recent success of deep learning has given rise to powerful forecasting models \cite{lstnet, informer, itransformer, patchtst, xpatch, dlinear}. However, the conventional offline training paradigm, which relies on a static, finite dataset, is fundamentally misaligned with real-world scenarios. In practice, time series data often arrives as a continuous stream, with new observations recorded at a high temporal frequency, such as every second or minute. Since offline methodologies are unsuitable for learning from such infinitely accumulating data, \emph{Online Time Series Forecasting (OTSF)}, a paradigm for incrementally updating models, has emerged for these real-world applications.

The primary challenge in OTSF is adapting to \emph{concept drift} \cite{cd}. In the context of OTSF, concept drift is the phenomenon where the underlying pattern of the data evolves over time. This evolution causes a mismatch between a model's historical training distribution and the continuously changing test distribution, inevitably leading to a drop in forecasting performance. While effectively addressing this challenge is critical for OTSF, the prevailing approaches \cite{dsof, proceed, fsnet, onenet} have been to treat concept drift as monolithic: they typically propose a single, unified adaptation strategy designed to react to any drift in the data stream. To address this oversimplification, our work begins by redefining the problem, categorizing drift into two distinct forms. The first is \textbf{Recurring Drift}, characterized by the reappearance of transitional patterns that are highly similar to the drifts observed in earlier time intervals. For instance, the drift in traffic volume from a Friday to a Saturday is statistically closer to the transition observed at the same weekly interval (e.g., Friday-to-Saturday of the last week) than to the immediately preceding Thursday-to-Friday shift. The second is \textbf{Emergent Drift}, where entirely new patterns arise from previously unobserved events, such as lifestyle changes from a pandemic altering electricity consumption, or the opening of a new public transportation fundamentally changing local traffic flow.

A closer examination of existing works reveals a critical weakness in handling Recurring Drift. Most of the current state-of-the-art methods employ a dual structure: a general-purpose backbone for stable learning, and an auxiliary mechanism that adapts the model with a strong focus on the most recent data \cite{fsnet, dsof, proceed}. However, we question whether this "recency-first" strategy is capable of mitigating \emph{Recurring Drift}, where relevant information may lie further in the past. Indeed, as our empirical analysis shows (Section~\ref{empirical}), this recency-focused strategy proves particularly ineffective when \emph{recurring patterns} become more relevant than \emph{immediate past}.

To this end, we propose \textbf{DynaME} (Dynamic Multi-period Experts), a novel framework built on a hybrid architecture that combines a stable, general-purpose backbone with a committee of highly adaptive, specialized experts. The committee of experts is designed to explicitly tackle \emph{Recurring Drift}. At each time step, the framework analyzes the current context to identify a variety of dominant periodic patterns across different time scales (e.g., daily, weekly). It then assigns a dedicated expert to each of these periods, instantly specializing them by rapidly fitting on the relevant historical data for that specific time scale. Complementing the expert committee, our Dynamic Gating Network serves as the central coordinator, intelligently weighing the predictions from the experts to determine which periodic patterns are most influential for the current forecast. This dynamic, per-step specialization allows the model to adaptively leverage the most relevant recurring patterns.

In addition, our framework is also designed to be effective against \emph{Emergent Drift}. The Dynamic Gating Network is equipped with a safety mechanism that monitors for a "danger signal", which quantifies sharp spikes in recent prediction error and indicates a situation where the specialized experts often become unreliable. In response, the gate shifts its reliance to a parameterized \emph{generalized expert} included within the committee. This provides a robust forecast to prevent large prediction errors, while at the same time leveraging this expert's plasticity to learn the new pattern. This generalized expert thus serves as a stabilizing buffer during the online process, ensuring robustness until the specialized experts can re-adapt to the newly observed historical data.

Through extensive experiments on several benchmark datasets, we demonstrate that DynaME significantly outperforms existing OTSF methods. We validate the general applicability of our framework across various state-of-the-art backbone architectures and provide comprehensive analyses, including computational efficiency and in-depth ablation studies, to justify our key design choices.

Our contributions are summarized as follows:
\begin{itemize}
    \item We redefine concept drift in Online Time Series Forecasting (OTSF) by categorizing it into two distinct types: \emph{Recurring Drift} and \emph{Emergent Drift}, and provide an empirical analysis that reveals the limitations of recency-focused methods in handling Recurring Drift.
    \item We propose \textbf{DynaME}, a hybrid framework specifically designed to handle both drift types by combining a backbone with a dynamic committee of experts.
    \item Through extensive experiments, we demonstrate that DynaME significantly outperforms state-of-the-art methods across various benchmark datasets and backbones.
\end{itemize}

\section{Preliminaries}

\subsection{Time Series Forecasting}

A time series is a sequence of observations ordered in time. Let $\mathcal{S} = (\mathbf{s}_1, \mathbf{s}_2, \mathbf{s}_3,\dots)$ be a time series, where each observation $\mathbf{s}_t \in \mathbb{R}^{C}$ is a vector of $C$ variables or channels. The task of time series forecasting is to predict future values based on past observations. Specifically, at any given time step $t$, the goal is to predict the next $H$ future values, known as the horizon window. To do this, we use a fixed-length window of the most recent past observations, known as the lookback window. We define them as follows:
\begin{itemize}[leftmargin=2em]
    \item \textbf{Lookback Window}: $\textbf{x}_t = (\textbf{s}_{t-L+1}, \dots, \textbf{s}_t) \in \mathbb{R}^{L \times C}$, which consists of the last $L$ observations.
    \item \textbf{Horizon Window}: $\textbf{y}_t = (\textbf{s}_{t+1}, \dots, \textbf{s}_{t+H}) \in \mathbb{R}^{H \times C}$, which consists of the next $H$ future observations.
\end{itemize}
The forecasting task is to learn a model $\Phi$ that takes the input $\mathbf{x}_t$ to produce a prediction $\hat{\mathbf{y}}_t$, such that the error (e.g., Mean Squared Error) between $\hat{\mathbf{y}}_t$ and the true value $\mathbf{y}_t$ is minimized. 

\subsection{Online Time Series Forecasting}

The Online Time Series Forecasting (OTSF) paradigm fundamentally differs from conventional offline training. In this setting, data arrive as a continuous stream, and new observations must be processed sequentially.
Here, we denote the complete state of $\Phi$ as $\Theta$, including not only the model parameters but also auxiliary components such as the experience replay buffer commonly used \cite{er, fsnet, dsof, solid}, and any additional modules required for adaptation \cite{proceed, onenet, dsof}.
The process at any given time step $t$, after the new observation $\mathbf{s}_t$ has arrived, can be formally described by the following sequence of operations:

\begin{itemize}[leftmargin=2em]
    \item \textbf{Adaptation}: Before making a new prediction, the model first updates its state from the previous state $\Theta_{t-1}$ to the current state $\Theta_t$. This adaptation step, $\mathcal{A}$, leverages the new observation $\mathbf{s}_t$ to keep the model aligned with the evolving data stream. The adaptation is generally formulated as follows:
    $$ \Theta_{t} = \mathcal{A}(\mathbf{s}_t, \Theta_{t-1}). $$

    \item \textbf{Prediction}: With its updated state $\Theta_t$, the model observes the most recent input lookback window $\mathbf{x}_t$, and generates a forecast for the next $H$ steps, denoted as $\hat{\mathbf{y}}_t = \Phi_{\Theta_t}(\mathbf{x}_t)$.
\end{itemize}
This cycle of updating with new information and then predicting is repeated for the entire duration of the online data stream, simulating a model's ability to learn and adapt in the evolving environment.

\section{Empirical Analysis}\label{empirical}
In this work, we redefine concept drift into \emph{Recurring Drift}, where previously seen patterns reappear, and \emph{Emergent Drift}, where entirely unforeseen patterns emerge. From this new perspective, this section provides an empirical analysis that highlights the limitations of the "recency bias" inherent in existing OTSF methods and underscores the necessity of a dynamic adaptation mechanism.

\subsection{The Limitation of Recency-Biased Adaptation}

Recent advanced methods have largely succeeded by prioritizing adaptation based on the most recent data \cite{proceed, dsof}. We hypothesize that this focus on recency is not universally effective, as it may overlook crucial long-term historical patterns relevant to Recurring Drift. To investigate this, we conducted a comparative experiment on the Traffic dataset \cite{lstnet}, comparing how PROCEED \cite{proceed}, a state-of-the-art recency-focused method, and a simple seasonal baseline, \emph{Weekly-LR}, react to Recurring Drift. \emph{Weekly-LR} is a simple linear regression model we introduce for this analysis, which is trained solely on four historical data points sampled at weekly intervals.

\begin{figure}[]
    \centering
    \includegraphics[width=1.0\linewidth]{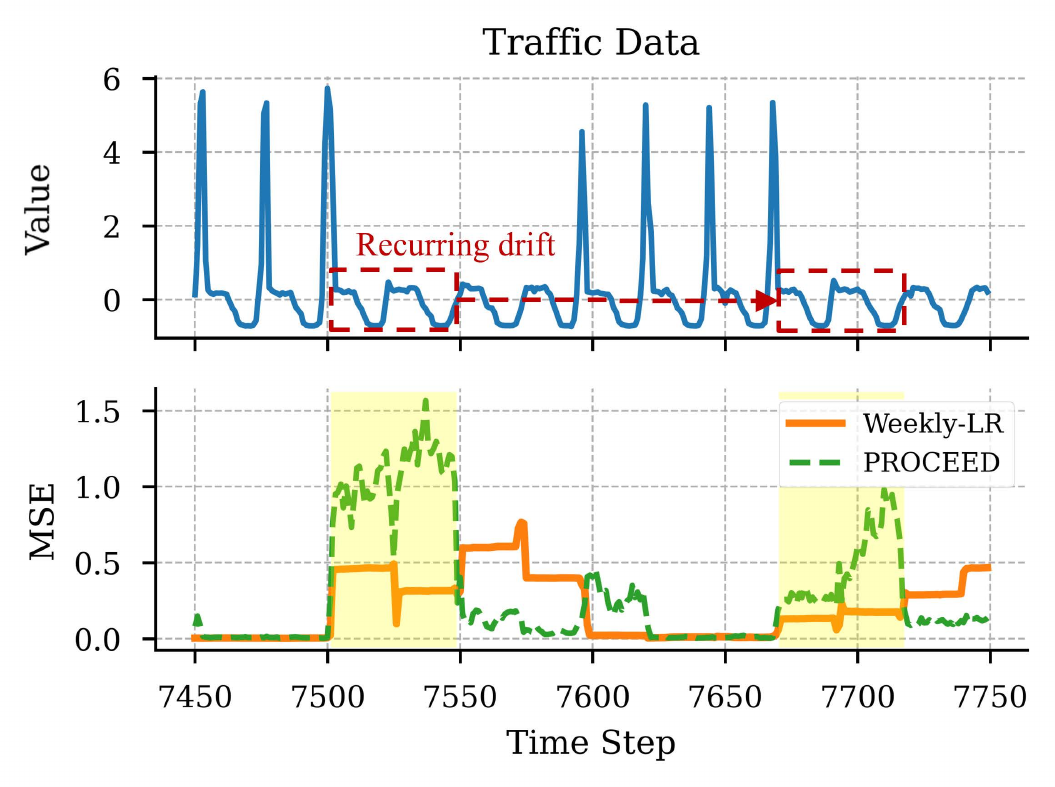}
    \caption{Effect of recurring drift on PROCEED and Weekly-LR performance on the Traffic dataset, showing the time series segment (top) and corresponding MSE (bottom).}    \label{weeklylr}
\end{figure}

The results in Figure~\ref{weeklylr} highlights a weakness in recency-first adaptation strategies. We observed that while generally effective, PROCEED's performance could fall behind the simple Weekly-LR model in specific moments dominated by Recurring Drift. In Figure~\ref{weeklylr}, a clear example occurs at timesteps 7500 and 7700, where the time series shifts to a different, but previously observed recurring pattern. PROCEED, with its recency-focused adaptation, attempts to predict this change based on the immediately preceding drifts, leading to a large prediction error. In contrast, Weekly-LR leverages the proper historical patterns to accurately predict the transition, thus achieving a lower error. This reveals that an adaptation strategy focused on recency risks forgetting relevant historical patterns.

\subsection{The Dynamic Nature of Periodicity}

The analysis in Section 3.1 establishes that for Recurring Drift, leveraging the correct historical context is more critical than relying on a simple recency bias. However, this leads to a more complex question: how should we determine which part of the past is the most relevant for the current prediction? We hypothesize that the most informative historical periodicity is not a static property, but a dynamic variable that changes depending on both the dataset's characteristics and the local context of the data over time.

\begin{figure}[t]
    \centering
    \includegraphics[width=1.0\linewidth]{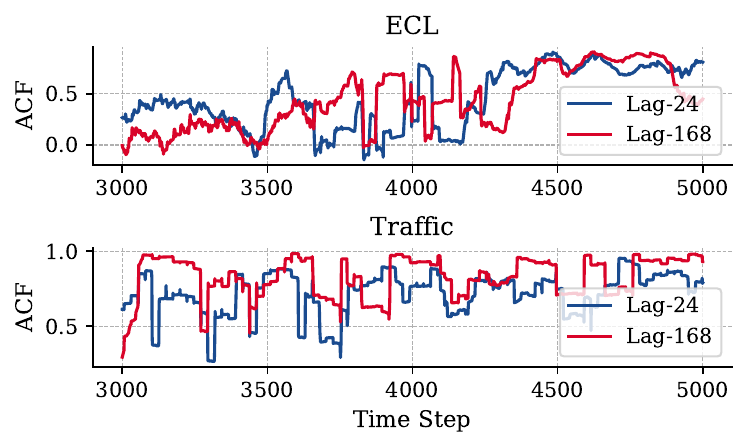}
    \caption{Autocorrelation Function (ACF) analysis on the ECL (top) and Traffic (bottom) datasets. Each plot compares the correlation strength of the daily (Lag-24) and weekly (Lag-168) patterns.}
    \label{acl}
\end{figure}

To investigate this, we analyzed the Autocorrelation Function (ACF) of the ECL and Traffic datasets \cite{lstnet}. The ACF measures the similarity of a time series with a lagged version of itself, providing a clear indicator of which historical periods are most relevant to the present. For instance, if the ACF value at Lag-168 is higher than at Lag-24, it implies that the data from one week ago is a more relevant predictor for the current data than the data from yesterday. Our findings, illustrated in Figure~\ref{acl}, validate our hypothesis by showing that the lag with the highest autocorrelation is not constant. For the ECL dataset, the dominant lag frequently alternates between Lag-24 and Lag-168. In the Traffic dataset, the correlation at Lag-168 is predominantly higher than at Lag-24, but there are still distinct intervals where the daily correlation becomes more significant. This evidence necessitates a mechanism that can dynamically identify and prioritize the most relevant periodicity at each step.

\section{Proposed Method: DynaME}

Motivated by the empirical analysis, we propose \textbf{DynaME} (Dynamic Multi-period Experts), a framework that makes predictions using a stable backbone and a committee of experts specialized for diverse periods. In this section, we describe the overall architecture and the step-by-step process of adaptation and prediction.

\subsection{Overall Architecture}
\begin{figure*}
    \centering
    \includegraphics[width=1.0\linewidth]{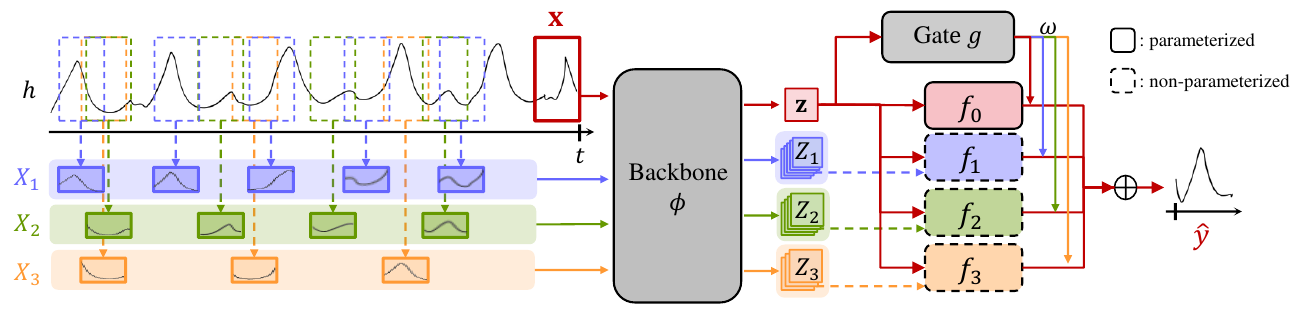}
    \caption{DynaME architecture. The solid-bordered components ($\phi$, $f_0$, $g$) are parameterized and trained, while the dash-bordered components ($f_1$, $f_2$, $f_3$) are non-parametrically fitted.}

    \label{main}
\end{figure*}

The architecture of DynaME, depicted in Figure \ref{main}, consists of a shared backbone, a committee of expert predictors, and the Dynamic Gating Network. The process unfolds in the following steps:
\begin{enumerate}[leftmargin=2em]
\item \textbf{Dynamic Period Analysis}: It begins by identifying the most dominant periods from the current context. For each period, a small specialized batch of relevant historical data is prepared.
\item \textbf{Representation Extraction}: The shared backbone then processes both the current input and the prepared batches to extract high-level feature representations.
\item \textbf{Expert Prediction}: These representations are then used for prediction. The backbone's standard prediction head acts as a generalist expert and makes a forecast. Simultaneously, each specialized expert is rapidly adapted on its dedicated historical batch and generates its own specialized forecast.
\item \textbf{Dynamic Gating}: The representation of the current input is also fed into the Dynamic Gating Network. This network assesses the current context and outputs a set of weights, indicating how reliable each expert's forecast is likely to be.
\item \textbf{Final Prediction}: Finally, the model's ultimate forecast is produced by taking a weighted average of all the predictions from the generalized and specialized experts, using the weights provided by the gating network.
\end{enumerate}

\subsection{Dynamic Period Selection for Experts}

The framework dynamically determines the most relevant periodicities from the current history window $\mathbf{h}_t$. This history window is a buffer containing the most recent $M$ observations, defined as:
$$ \mathbf{h}_t = (v_{t-M+1}, \dots, v_t) \in \mathbb{R}^{M \times C}. $$
We apply the Fast Fourier Transform (FFT) to this history window, from which we select the set of top-$k$ dominant frequencies, $\mathcal{F}$, as follows \cite{timesnet, perimid}:
\begin{equation}
  \mathcal{F} = \underset{i \in \{1, \dots, \lfloor M/2 \rfloor \}}{\text{arg top-k}} \left( \frac{1}{C} \sum_{c=1}^{C} \left|\text{FFT}(\mathbf{h}^c)[i]\right| \right),
\end{equation}
where $\mathbf{h}^c$ is the sequence for the $c$-th channel and $|\cdot|$ denotes the amplitude of the frequency component\footnote{For notational simplicity, we will omit the time subscript $t$. Unless otherwise specified, all variables refer to the current time step $t$.}. These frequencies are then converted into a set of period lengths, $\mathcal{P} = \{ \lfloor L/i \rfloor \mid i \in \mathcal{F} \}$.

Once the set of dominant period lengths $\mathcal{P}$ is identified, the next step is to construct a historical data batch for each of the $k$ specialized experts. To provide a consistent role for each expert (e.g., long-term vs. short-term), the periods are first sorted by their length in descending order, yielding an ordered list $[p_1, p_2, \dots, p_k]$. For each sorted period $p_i$, we construct a corresponding input batch $X_i \in \mathbb{R}^{n_i \times L \times C}$ and a target batch $Y_i \in \mathbb{R}^{n_i \times H \times C}$. These batches are formed by sampling $n_i$ historical input-output pairs from $\mathbf{h}$ at intervals of the period length $p_i$. For instance, if a weekly period ($p=168$) is identified, the expert's training batch is constructed using data from one week ago ($t-168$), two weeks ago ($t-336$), and so on. Each specialized expert is then assigned one of these period-specific data batches for its training.

\subsection{Non-parametric Expert Adaptation}

This committee is composed of two distinct types of predictors. The first, $f_0$, is a parameterized \emph{generalized expert}, providing a stable, generalized forecast. In contrast, the other $k$ experts, $\{f_1, \dots, f_k\}$, are \emph{specialized experts} that are highly adaptive. They must be fitted on-the-fly at each time step, each on its own designated batch of period-specific historical data. This requirement for per-step adaptation would present a significant computational challenge if these experts were to be fine-tuned with standard gradient descent.

However, by interpreting the specialized expert's task as a linear regression problem, we can bypass iterative gradient descent and solve for the optimal weights directly. Let $Z_i = \phi(X_i)\in \mathbb{R}^{n_i \times D}$ be the features from the $i$th expert's assigned batch and $Y_i \in \mathbb{R}^{n_i \times H}$ be the corresponding targets\footnote{Since the regression for each expert is performed independently per channel, we formulate the problem for a single channel for notational simplicity.}. The expert's optimal weight matrix $W_i \in \mathbb{R}^{D \times H}$ can be obtained by minimizing the Ridge Regression objective:
\begin{equation}
    \min_{W_i} \left\{ \| Y_i - Z_i W_i \|_F^2 + \lambda \|W_i\|_F^2 \right\}.
\end{equation}
While the direct closed-form solution is $W_i= (Z_i^T Z_i + \lambda I)^{-1} Z_i^T Y_i$, it has a computational bottleneck of $O(D^3)$ due to the inversion of the $D \times D$ matrix. Given that the representation dimension $D$ is typically large, this is expensive to perform at every step.

To overcome this bottleneck, we reformulate the problem to solve it in its dual form \cite{krr}. This approach shifts the computational complexity from the feature dimension $D$ to the number of fitting samples $n$. Let $\mathbf{z} = \phi(x) \in \mathbb{R}^{D}$ be the feature representation for the current input. The prediction $\hat{\mathbf{y}}_i$ is then formulated directly without computing $W_i$:
\begin{equation}
    \hat{\mathbf{y}}_i = (\mathbf{z} Z_i^T) (Z_i Z_i^T + \lambda I)^{-1} Y_i.
\end{equation}
The dominant computation is now the inversion of the $n_i \times n_i$ matrix $(Z_i Z_i^T+\lambda I)$, resulting in a complexity of $O(n_i^3)$. By using a small number of samples for fitting ($n_i \ll D$), this process becomes computationally feasible.
\subsection{Final Prediction with Dynamic Gating}\label{sec:final}

The final prediction is a weighted combination of all expert outputs. The initial weights, $\tilde{\omega} \in \mathbb{R}^{k+1}$, are determined by the dynamic gating network, a mapping function $g: \mathbb{R}^{D} \to \mathbb{R}^{k+1}$. This network takes the $D$-dimensional feature representation $z$ as input and is implemented as a 2-layer MLP with a GELU \cite{gelu} activation function. The process can be described as:
\begin{equation}
    \tilde{\omega} = \text{Softmax}(\text{MLP}_{\text{out}}(\text{GELU}(\text{MLP}_{\text{in}}(z)))),
\end{equation}
where $\text{MLP}_{\text{in}}$ projects the input to a hidden space, and $\text{MLP}_{\text{out}}$ maps it to the final $k+1$ dimensional output for the expert weights.

However, this weighting scheme assumes that future patterns can be represented as a combination of known historical patterns, which may fail under \emph{Emergent Drift}.
When entirely new patterns arise, the historical data used to fit the specialized experts becomes irrelevant, leading to sharply increased prediction errors. In such cases, the model must ensure stability while enabling rapid adaptation. To address this, we propose a mechanism that adaptively adjusts the weight $\tilde{\omega}$ based on a danger signal $d_t$ derived from recent prediction errors. This signal identifies when specialized experts are likely failing, prompting a shift in reliance toward the model’s robust component: the parameterized, generalized expert $f_0$. This allows the model to fall back on a stable forecast to suppress error spikes, while simultaneously focusing the learning pressure onto $f_0$, compelling it to rapidly adapt to the new pattern. In this situation, the generalized expert acts as a buffer until the specialized experts acquire sufficient data to recover performance.

This process begins by tracking the most recent MSE at step $t$, denoted $\text{MSE}_t$, using an Exponentially Weighted Moving Average (EWMA) $\mu_t^{\text{MSE}}$ as follows:
\begin{equation}
    \mu_t^{\text{MSE}} = (1 - \alpha) \cdot \text{MSE}_t + \alpha \cdot \mu_{t-1}^{\text{MSE}},
\end{equation}
where $\alpha$ is a smoothing factor. A large deviation of the current MSE from its EWMA is interpreted as a signal of an unforeseen event. This deviation is transformed into a normalized danger signal, $d_t \in [0, 1]$, using a non-linear sensitivity function:
\begin{equation}
    d_t = 1 - \exp(-\delta (\text{MSE}_t - \mu_t^{\text{MSE}})^2).
\end{equation}
This allows the danger signal to remain insensitive to subtle fluctuations (e.g., between 0 and 1) while reacting sharply to large spikes (e.g., deviations over 10). Finally, the initial weights $\tilde{\omega}$ are corrected by blending them with a backbone-only weight vector to produce the final weights $\omega$. This process is guided by the blending factor $\gamma$, which is dynamically mapped from the danger signal $d_t$. The entire correction is formulated as follows:

\begin{equation}
\begin{aligned}
    \omega &= (1 - \gamma) \tilde{\omega} + \gamma [1, 0, \dots, 0], \\
    &\text{where} \quad \gamma = \beta + d_t (1 - \beta).
\end{aligned}
\end{equation} Here, $\beta$ is a hyperparameter representing a minimum reliance on the backbone. The final prediction $\hat{y}$ is then computed using these weights as $\hat{\mathbf{y}} = \sum_{i=0}^{k} \omega_{i} \cdot \hat{\mathbf{y}}_{i}$. This mechanism allows our model to operate in its high-performance, multi-expert mode during stable periods (low $d_t \Rightarrow \gamma \approx \beta$) but gracefully fall back to $f_0$ during abrupt changes (high $d_t \Rightarrow \gamma \approx 1$). 

The entire framework is trained end-to-end. Once the ground truth $\mathbf{y}$ for a given prediction $\hat{\mathbf{y}}$ is fully observed, a standard MSE loss is computed as 
$\mathcal{L}=\text{MSE}(\hat{\mathbf{y}}, \mathbf{y})$. The resulting gradient $\nabla\mathcal{L}$ is then backpropagated to update the trainable parameters of both the backbone $\phi$, the general expert $f_0$, and the Dynamic Gating Network $g$. A crucial consideration in this process is the inherent \emph{feedback delay} \cite{dsof, proceed}. To respect the online setting and prevent information leakage, the update must only use fully observed data. Therefore, the adaptation performed at the current time step $t$ is not based on the prediction just made ($\hat{y}_t$), but on the most recent prediction for which the ground truth has become complete, which is the one made at step $t-H$. For an overview of the adaptation and prediction process, refer to Algorithm~\ref{alg:forward} and Algorithm~\ref{alg:online} in Appendix~\ref{app:implement}.

\begin{table*}[t]
\centering
\caption{Comparison of methods across different backbones and prediction horizons. Performance is measured in \textbf{MSE} (Mean Squared Error), with the best performance in each column highlighted in bold.}
\label{main_table}
 \renewcommand{\arraystretch}{0.95}
\resizebox{\textwidth}{!}{
\begin{tabular}{clccccccccccccccc}
    \toprule
     & \textbf{Dataset} & \multicolumn{3}{c}{ETTh2} & \multicolumn{3}{c}{ETTm1} & \multicolumn{3}{c}{Traffic} & \multicolumn{3}{c}{ECL} & \multicolumn{3}{c}{Weather} \\
    \cmidrule(lr){3-5}\cmidrule(lr){6-8}\cmidrule(lr){9-11}\cmidrule(lr){12-14}\cmidrule(lr){15-17}
     & \textbf{Horizon ($H$)} & $24$ & $48$ & $96$ & $24$ & $48$ & $96$ & $24$ & $48$ & $96$ & $24$ & $48$ & $96$ & $24$ & $48$ & $96$ \\
    \midrule
    
    \multirow{5}{*}{\rotatebox{90}{PatchTST}} 
    & GD & 2.062 & 3.412 & 5.838 & 0.524 & 0.675 & 0.763 & 0.511 & 0.541 & 0.533 & 4.225 & 4.926 & 6.096 & 0.927 & 1.304 & 1.589 \\
    & SOLID & 1.924 & 3.439 & 5.801 & 0.498 & 0.645 & 0.779 & 0.403 & 0.557 & 0.544 & 4.191 & 4.897 & 6.122 & 0.897 & 1.226 & 1.612 \\
    & DSOF & 1.850 & 3.591 & 6.242 & 0.595 & 0.822 & 0.950 & 0.478 & 0.536 & 0.546 & 6.363 & 7.742 & 8.441 & 0.998 & 1.532 & 1.920 \\
    & PROCEED & 1.711 & 2.905 & 5.752 & 0.494 & 0.659 & 0.776 & 0.395 & 0.435 & 0.457 & 4.153 & 4.705 & 5.871 & 0.855 & 1.212 & 1.578 \\
    & DynaME (ours) & \textbf{1.676} & \textbf{2.871} & \textbf{5.049} & \textbf{0.439} & \textbf{0.589} & \textbf{0.703} & \textbf{0.376} & \textbf{0.398} & \textbf{0.431} & \textbf{4.073} & \textbf{4.541} & \textbf{5.837} & \textbf{0.845} & \textbf{1.146} & \textbf{1.456} \\
    \midrule
    
    \multirow{5}{*}{\rotatebox{90}{iTransformer}} 
    & GD & 1.934 & 3.134 & 5.408 & 0.501 & 0.658 & 0.742 & 0.392 & 0.427 & 0.439 & 3.899 & 4.653 & 5.869 & 0.891 & 1.228 & 1.556 \\
    & SOLID & 1.991 & 3.202 & 5.305 & 0.486 & 0.640 & 0.726 & 0.423 & 0.467 & 0.430 & 3.952 & 4.726 & 5.957 & 0.910 & 1.258 & 1.581 \\
    & DSOF & 1.989 & 3.378 & 5.943 & 0.630 & 0.880 & 1.061 & 0.422 & 0.460 & 0.479 & 5.127 & 6.348 & 8.231 & 1.023 & 1.430 & 1.813 \\
    & PROCEED & 1.968 & 3.255 & 6.182 & 0.473 & 0.645 & 0.754 & 0.402 & 0.434 & 0.429 & 3.857 & 4.585 & 5.708 & 0.834 & 1.165 & 1.475 \\
    & DynaME (ours) & \textbf{1.854} & \textbf{2.946} & \textbf{5.059} & \textbf{0.465} & \textbf{0.598} & \textbf{0.682} & \textbf{0.395} & \textbf{0.413} & \textbf{0.417} & \textbf{3.834} & \textbf{4.574} & \textbf{5.685} & \textbf{0.828} & \textbf{1.132} & \textbf{1.436} \\
    \midrule

    \multirow{5}{*}{\rotatebox{90}{xPatch}} 
    & GD & 1.895 & 3.431 & 5.681 & 0.598 & 0.748 & 0.811 & 0.492 & 0.519 & 0.505 & 4.329 & 5.009 & 6.167 & 0.982 & 1.311 & 1.588 \\
    & SOLID & 2.249 & 3.428 & 5.260 & 0.502 & 0.650 & 0.734 & 0.474 & 0.503 & 0.494 & 4.338 & 5.039 & 6.183 & 1.014 & 1.344 & 1.631 \\
    & DSOF & 1.883 & 3.373 & 6.020 & 0.564 & 0.798 & 0.957 & 0.516 & 0.550 & 0.533 & 5.096 & 6.232 & 8.755 & 0.863 & 1.306 & 1.816 \\
    & PROCEED & 2.063 & 4.032 & 6.675 & 0.557 & 0.711 & 0.788 & 0.459 & 0.499 & 0.498 & 4.272 & 4.897 & 5.984 & 0.892 & 1.241 & 1.519 \\
    & DynaME (ours) & \textbf{1.845} & \textbf{3.139} & \textbf{5.206} & \textbf{0.469} & \textbf{0.601} & \textbf{0.678} & \textbf{0.417} & \textbf{0.434} & \textbf{0.442} & \textbf{4.177} & \textbf{4.828} & \textbf{5.922} & \textbf{0.829} & \textbf{1.108} & \textbf{1.401} \\
    \bottomrule
\end{tabular}
 \renewcommand{\arraystretch}{1.0}
}
\Description{The table presents the Mean Squared Error (MSE) for five different time series forecasting methods (GD, SOLID, DSOF, PROCEED, DynaME (ours)) using three different transformer backbones (PatchTST, iTransformer, xPatch). The best (lowest) MSE value in each combination of dataset and horizon is highlighted in bold.}
\end{table*}

\section{Experiments}
In this section, we conduct a series of experiments to answer the following key questions: (RQ1) Does DynaME outperform state-of-the-art online forecasting methods? (RQ2) Is our proposed framework computationally efficient for real-world scenarios? (RQ3) Are the core components of DynaME essential for its performance? 

\subsection{Settings}

Before presenting our results, we first outline the experimental settings, including the datasets, backbone architectures, and baseline methods used for comparison.

\smallsection{Datasets} We evaluate our model on widely-used benchmark datasets, including ETT, Traffic, ECL, and Weather datasets \cite{lstnet, informer}. Following the conventional protocol for OTSF, we pre-train our models on the initial 25\% of each dataset (20\% for training and 5\% for validation) and simulate the online test phase on the remaining 75\%. For the details, please refer to Appendix~\ref{app:dataset}.

\smallsection{Backbones} To demonstrate the robustness and general applicability of our framework, we implement DynaME on top of several widely used state-of-the-art backbone models:
\begin{itemize}[leftmargin=2em]
    \item \textbf{PatchTST} \cite{patchtst}: A Transformer-based model that treats patches of a time series as input tokens.
    \item \textbf{iTransformer} \cite{itransformer}: An inverted Transformer architecture that processes individual time series as tokens to better capture cross-variable correlations.
    \item \textbf{xPatch} \cite{xpatch}: A  dual-stream MLP \& CNN architecture that utilizes patching and exponential decomposition.
\end{itemize}

\smallsection{Baselines} We compare DynaME against a comprehensive suite of baselines for OTSF, ranging from simple methods to recent, state-of-the-art frameworks:
\begin{itemize}[leftmargin=2em]
    \item \textbf{GD}: A simple baseline that continuously fine-tunes the entire model on newly arriving data using gradient descent.
    \item \textbf{DSOF} \cite{dsof}: A framework that uses fast and slow streams for online updates.
    \item \textbf{PROCEED} \cite{proceed}: A proactive adaptation method that models the drift caused by feedback delays.
    \item \textbf{SOLID} \cite{solid}: An adaptation framework that fine-tunes the model on a small, curated batch of historical data that is contextually similar to the current input.
\end{itemize}
In all of our experiments, we strictly adhere to a realistic online setting that considers \emph{feedback delay}. Specifically, we ensure that information leakage does not occur when updating the model parameters or $d_t$ with $\text{MSE}_t$. Methods that do not follow this constraint are not included in our direct comparison to maintain fairness and practical relevance.

\subsection{Performance Analysis (RQ1)}

We evaluate the performance of DynaME against the baseline methods across five benchmark datasets, three different backbone architectures, and three forecast horizons. The comprehensive results, measured in Mean Squared Error (MSE), are presented in Table~\ref{main_table}. We discuss the key insights from these results below.

\smallsection{Consistent State-of-the-Art Performance}
Across all datasets and forecast horizons, DynaME sets a new state-of-the-art for online time series forecasting. It consistently outperforms recent, strong recency-based adaptation methods like DSOF and PROCEED. Moreover, the consistent outperformance over SOLID reveals that a single fixed strategy is insufficient to fully address Recurring Drift. This confirms our hypothesis: by explicitly redefining concept drift and addressing both its Recurring and Emergent drifts with specialized mechanisms, our framework provides a more comprehensive solution than methods that rely on a single, monolithic adaptation strategy.

\smallsection{Model-Agnostic Generalization}
A key strength of DynaME is its general applicability. The performance gains are not limited to a single backbone but are consistently observed across all three state-of-the-art architectures: PatchTST, iTransformer, and xPatch. This demonstrates that our framework is a model-agnostic plug-in that effectively enhances the online adaptation capabilities of various advanced forecasting models.

\smallsection{Robustness Across Forecast Horizons}
The results also indicate that DynaME's advantage is maintained or even amplified as the forecast horizon increases. In a majority of cases, the relative improvement of DynaME over the baselines becomes larger for longer horizons (e.g., $T$=96). As longer horizons naturally demand a greater reliance on long-term patterns over recent ones, our framework’s effective utilization of relevant history provides a clear advantage.

\subsection{Efficiency Analysis (RQ2)}

As our framework utilizes a committee of experts, concerns may arise regarding potential drawbacks in computational efficiency. In this section, we analyze DynaME's performance from two key aspects: memory overhead and inference time.

\smallsection{Memory Overhead} A potential concern with an expert committee architecture is the memory overhead required to store multiple models. In DynaME, however, the design is inherently parameter-efficient. Beyond the backbone itself, which can contain up to approximately 10M parameters \cite{itransformer, patchtst, dlinear}, the only additional trainable parameters reside in the lightweight Dynamic Gating Network, adding merely up to 0.1M parameters. Furthermore, the specialized experts are non-parametric, which add no permanent parameter overhead. Consequently, the total memory footprint of DynaME remains comparable to that of a single backbone model.

\smallsection{Inference Time}
The inference time is also carefully managed to avoid significant latency. The primary computational cost comes from fitting the specialized experts at each step. By solving this fitting problem in its dual form, we shift the main computational complexity from the high feature dimension $D$ to the small number of fitting samples $n$. This reduces the complexity from a prohibitive $O(D^3)$ to a much more feasible $O(n^3)$, as a small number of samples (e.g. $4\sim 8$) are used for fitting.

To validate the real-world impact of our design, we measured average wall-clock time, comparing our standard model (DynaME-dual) with a primal-form variant (DynaME-primal) and other baselines. Results in Table~\ref{tab:inference} show that DynaME-primal is significantly slower, confirming the prohibitive cost of a naive implementation, while DynaME-dual is substantially faster, demonstrating that the dual-form solution is essential for practical on-the-fly adaptation. Although DynaME-dual incurs higher computational cost than simpler methods like GD due to its sophisticated specialization mechanism, its speed remains comparable to advanced baselines such as PROCEED. This confirms that DynaME achieves substantial accuracy gains at a practical computational cost.
\subsection{Ablation Study (RQ3)}

We conduct a series of ablation studies to validate the design choices of our key components: the Dynamic Gating Network and its integrated safety mechanism with the danger signal.

\subsubsection{Effectiveness of Dynamic Gating Network}
To verify the importance of our Dynamic Gating Network, we compare our full model against several naive alternatives:
\begin{itemize}[leftmargin=2em]
    \item \textbf{Simple Average}: Replaces the gating network with a simple, non-weighted average of all expert predictions.
    \item \textbf{Learnable Gate}: Uses a learnable set of weights which is independent of the input $\mathbf{x}$.
    \item \textbf{Detached Gate}: Replaces $\mathbf{z}$ with the raw data $\mathbf{x}$ as the gate's input, thus preventing the joint optimization of $g$ and $\phi$.
    
\end{itemize}
As shown in Table~\ref{tab:ablation_gate}, all simplified variants perform worse than the original Dynamic Gating Network. This demonstrates that the ability to dynamically weigh experts based on the current context is crucial for performance. The performance degradation of the Detached Gate also demonstrates that the gate needs the backbone's refined features ($\mathbf{z}$) to function, underscoring the necessity of their joint optimization.

\begin{table}[]
\caption{Average wall-clock time per step.} \label{tab:inference}
\renewcommand{\arraystretch}{0.9}
\begin{tabular}{lccc}
\toprule
Method & PatchTST & iTransformer & xPatch \\
\midrule
GD                   & 0.0277   & 0.0174       & 0.0082 \\
SOLID                & 0.0283   & 0.0176       & 0.0084 \\
DSOF                 & 0.0321   & 0.0255       & 0.0163 \\
PROCEED             & 0.0422   & 0.0367       & 0.0234 \\
\midrule
DynaME-primal        & 0.1252   & 0.1092       & 0.0754 \\
DynaME-dual (ours)        & 0.0489   & 0.0423       & 0.0279 \\
\bottomrule
\end{tabular}
\renewcommand{\arraystretch}{1.0}
\end{table}

\begin{table}[]
\caption{Performance comparison of different gating variants.}\label{tab:ablation_gate}
\renewcommand{\arraystretch}{0.97}
\resizebox{\columnwidth}{!}{
\begin{tabular}{lccccc}
\toprule
 & ETTh2 & ETTm1 & Traffic & ECL   & Weather \\
\midrule
Simple Average      & 3.001 & 0.467 & 0.407       & 4.962     & 1.005   \\
Learnable Gate       & 2.556 & 0.446 & 0.386       & 4.327    & 0.906   \\
Detached Gate       & 2.201     & 0.451 & 0.386       & 4.593     & 0.941   \\
Dynamic Gate (ours) & \textbf{1.676} & \textbf{0.439} & \textbf{0.376}   & \textbf{4.073} & \textbf{ 0.845}  \\
\bottomrule
\end{tabular}
\renewcommand{\arraystretch}{1.0}
}
\end{table}

\begin{table}[]
\caption{Performance comparison with and without $d_t$.}\label{ablation:dt}
\renewcommand{\arraystretch}{0.9}
\resizebox{\columnwidth}{!}{
\begin{tabular}{lcccccc}
\toprule
                  \textbf{Backbone}& \multicolumn{2}{c}{PatchTST} & \multicolumn{2}{c}{iTransformer} & \multicolumn{2}{c}{xPatch} \\
                  \cmidrule(lr){2-3}\cmidrule(lr){4-5}\cmidrule(lr){6-7}
                  \textbf{Dataset} & ETTh2      & Electricity     & ETTh2        & Electricity       & ETTh2     & Electricity    \\
                  \midrule
\texttt{w/o} $d_t$ & 1.906      & 4.905               & 2.318        & 4.719                 & 2.281     & 5.136              \\
\texttt{w/} $d_t$  & \textbf{1.676}      & \textbf{4.073}               & \textbf{1.854}        & \textbf{3.834}                 & \textbf{1.845}     & \textbf{4.177}           \\
\bottomrule
\end{tabular}}
\renewcommand{\arraystretch}{1.0}
\end{table}

\begin{figure}
    \centering
    \includegraphics[width=1.0\linewidth]{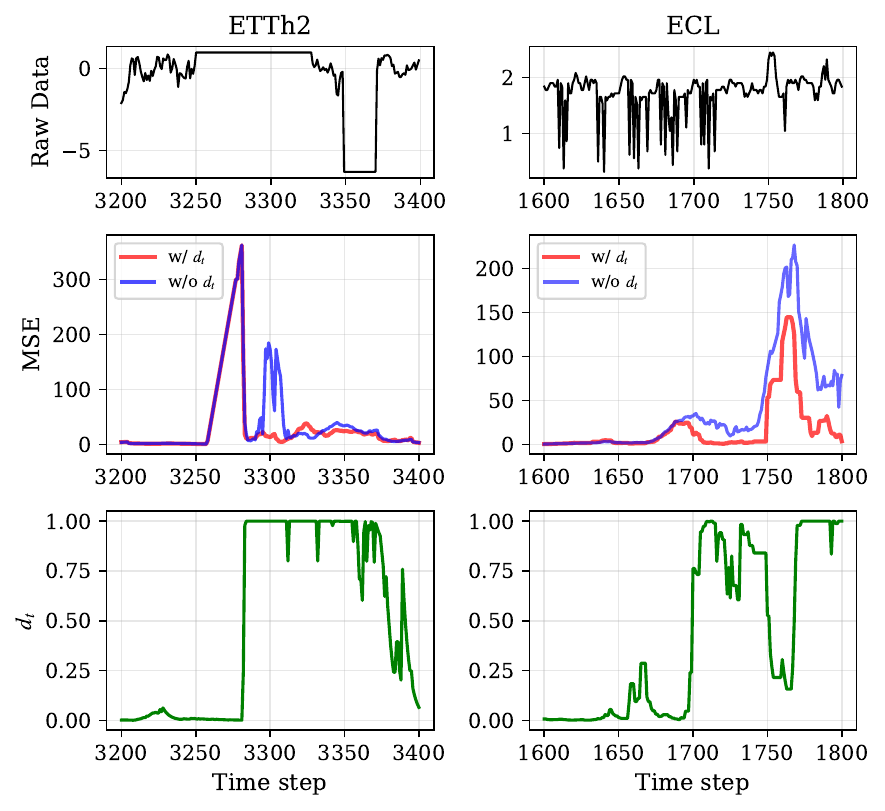}
    \caption{Per-step MSE of "\texttt{w/} $d_t$" and "\texttt{w/o} $d_t$" with danger signal $d_t$ during \emph{Emergent Drift}. The danger signal rises when an MSE spike is detected. }\label{fig:dt}
\end{figure}

\begin{table}[]
\caption{Performance comparison of dynamic vs. fixed period selection.}\label{tab:ablation_period}
\renewcommand{\arraystretch}{0.9}
\resizebox{\columnwidth}{!}{
\begin{tabular}{lccccc}
\toprule
\text{Period selection} & ETTh2 & ETTm1 & Traffic & ECL   & Weather \\
\midrule
Fixed      & 1.855 & 0.460 & 0.378       & 4.355     & 0.886  \\
Dynamic & \textbf{1.676} & \textbf{0.439} & \textbf{0.376}   & \textbf{4.073} & \textbf{ 0.845}  \\
\bottomrule
\end{tabular}
\renewcommand{\arraystretch}{1.0}
}
\end{table}

\subsubsection{Impact of the Danger Signal} \label{sec:danger}

To validate the effectiveness of our danger signal mechanism, we conducted an ablation study on the ETTh2 and ECL datasets, which contain volatile segments that produce sharp MSE spikes, a clear indicator of \emph{Emergent Drift}. We compare our full model ("\texttt{w/} $d_t$") against a variant where this mechanism is disabled ("\texttt{w/o} $d_t$"). While Table~\ref{ablation:dt} shows a significant overall performance drop without the danger signal, a deeper insight is gained from visualizing the per-step MSE and $d_t$ during these Emergent Drift (Figure~\ref{fig:dt}). In the volatile intervals (timesteps 3200-3400 for ETTh2 and 1600-1800 for ECL), it struggles to recover without $d_t$, as its MSE remains high and unstable. In contrast, our full model successfully detects the MSE spike as a danger signal, triggers the safety mechanism to shift reliance onto the stable backbone, and effectively suppresses the following errors. This analysis confirms that the danger signal is a critical component for ensuring stability and facilitating rapid adaptation during Emergent Drift.

\subsubsection{Effectiveness of Dynamic Period Selection}
To validate the effectiveness of our dynamic period selection mechanism, we compare our full method against a variant with fixed period selection. In this variant, we replace our FFT-based dynamic selection with a set of hard-coded, fixed periods based on well-known prior knowledge. Specifically, we assign the specialized experts to \emph{daily} and \emph{weekly} periods, which are dominant in a majority of time series datasets \cite{timesnet, fedformer, autoformer}. The results in Table~\ref{tab:ablation_period} demonstrate that our approach consistently outperforms the fixed setting. While the fixed approach performs reasonably well on the Traffic dataset, likely due to its strong daily/weekly periodicities (Figure~\ref{acl}), our method still achieves superior accuracy. This suggests that even in scenarios where clear periodic structures exist, relying solely on pre-defined seasonalities is suboptimal. Overall, the results confirm that dynamically identifying and adapting to the most influential periodicities is important for achieving robust forecasting performance.

\subsection{Hyperparameter Analysis}

\begin{figure}[t]
    \centering
    \includegraphics[width=1.0\linewidth]{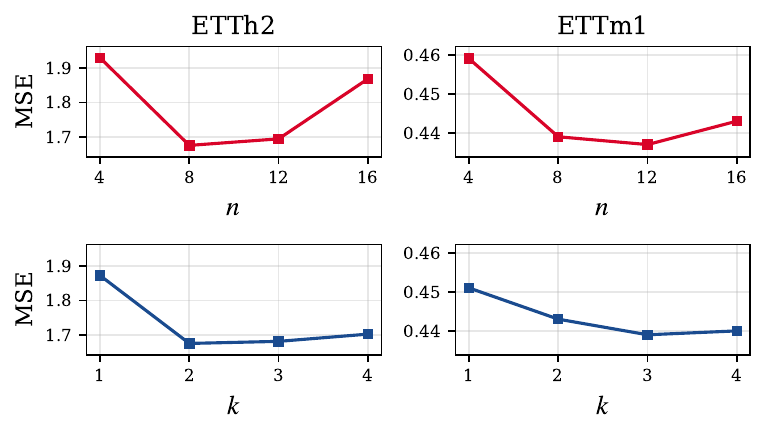}
    \caption{Performance comparison with varying Maximum Sample Size ($n$) and Number of Experts ($k$) on the ETTh2 and ETTm1 datasets.}\label{fig:hp}
\end{figure}

We analyze the sensitivity of DynaME to its two key hyperparameters: $n$, which governs the number of historical samples used for fitting each expert, and $k$, the number of specialized experts, Actually, the number of historical samples $n_i$ can vary for each expert. For simplicity, we control this with a single hyperparameter $n$ that defines the maximum number of samples for each experts ($n_i \le n$). The experiments were conducted on the ETTh2 and ETTm1 datasets, and the results are summarized in Figure~\ref{fig:hp}.

\smallsection{Analysis on Maximum Sample Size $n$}
As shown in the top half of Figure~\ref{fig:hp}, performance degrades when $n$ is small (e.g., 4), likely due to insufficient information. Conversely, performance also tends to degrade when $n$ is large (e.g., 16), indicating that incorporating excessively past data can introduce noise. For the ETT datasets, we find the optimal range for $n$ to be between 8 and 12. This suggests that effective adaptation hinges on referencing an optimal amount of the past, balancing the need for sufficient information against the risk of introducing noise from irrelevant historical data.

\smallsection{Analysis on Number of Experts $k$}
As shown in the bottom half of Figure~\ref{fig:hp}, using few experts (e.g., $k=1$) results in suboptimal performance, as the model lacks the capacity to capture multiple periodic patterns. While performance improves as more experts are added, it becomes stable and robust once a sufficient number are available. We attribute this robustness to the Dynamic Gating Network, which can learn to assign low weights to redundant or irrelevant experts, effectively filtering out unnecessary information during the final prediction.

\smallsection{Other Hyperparameters}
Our framework has other hyperparameters, such as those used in the safety mechanism ($\alpha, \delta$) or regression ($\lambda$). We found that the model's performance was robust to these parameters once they were set to reasonable values. Therefore, we used fixed values for these components throughout our experiments. For details of the setting, please refer to Appendix A.

\section{Related Work}
\smallsection{Time Series Forecasting}
The advancement of deep learning has led to significant performance improvements in Time Series Forecasting (TSF). The field has recently seen a surge in Transformer-based models that are effective in handling long-sequence data \cite{informer, patchtst, itransformer, perimid, fedformer, autoformer}. In parallel, simple MLP-based linear models have been proposed as an alternative to the complexity of Transformers, demonstrating surprising efficiency and strong performance \cite{dlinear, fits, sparsetsf}. Concurrently, it has been demonstrated that explicitly incorporating the inherent nature of time series, such as periodicity \cite{timesnet, perimid, sparsetsf, cyclenet, pdf} and seasonal-trend decomposition \cite{dlinear, xpatch, autoformer, timemixer}, into models is highly effective. In addition, several works have explored mixture of experts architectures for TSF to capture heterogeneous temporal patterns \cite{mole, timemoe}. Reflecting these trends, our work adopts PatchTST \cite{patchtst} and iTransformer \cite{itransformer} as Transformer-based backbones, and xPatch \cite{xpatch} as a MLP \& CNN-based backbone to demonstrate the generality of DynaME.

\smallsection{Online Time Series Forecasting}
The continuous, streaming nature of real-world time series data exposes the fundamental limitations of the conventional offline paradigm. Initial attempts to address this involved instance-wise adaptation to incoming inputs \cite{revin, nontf, dain} or the use of classic continual learning techniques \cite{cl, er,der}. However, recent research \cite{fsnet, onenet} has shifted focus towards tackling a more specific and prominent challenge in time series: concept drift, the phenomenon where underlying data patterns evolve over time. Prevailing research has tackled this challenge through several strategies, including rapid adaptation with a focus on recent data \cite{fsnet, proceed, dsof}, online ensembling methods \cite{onenet}, and context-aware fine-tuning \cite{solid}. However, these prevailing methods tend to treat concept drift as a single, monolithic phenomenon. In contrast, our work redefines concept drift into two distinct types: Recurring and Emergent Drift, and proposes DynaME, a framework specifically designed to effectively address both drifts.

\section{Conclusion}

In this work, we first revisited the notion of concept drift in OTSF and redefined it into two distinct forms: Recurring Drift and Emergent Drift. We then propose DynaME, addresses both drifts through a dynamic committee of generalized and specialized experts coordinated by Dynamic Gating Network.
Extensive experiments demonstrate that DynaME achieves consistent state-of-the-art performance while maintaining practical efficiency. As a promising future direction, extending DynaME to more expressive backbones such as foundation models could further enhance its adaptability and generalization to diverse real-world scenarios.


\begin{acks}
This work was supported by the National Research Foundation of Korea (NRF) grant funded by the Korea government (MSIT) (No. RS-2024-00335873), the Technology Innovation Program (No. RS-2025-02952974) funded by the Ministry of Trade, Industry \& Energy (MOTIE, Korea), Institute of Information \& communications Technology Planning \& Evaluation (IITP) grant funded by the Korea government(MSIT) (No.RS-2019-II191906, Artificial Intelligence Graduate School Program(POSTECH)), and the Korea Innovation Foundation (INNOPOLIS) grant funded by the Korea government (MSIT) (No. RS-2025-25449754).

\end{acks}

\bibliographystyle{ACM-Reference-Format}
\bibliography{dyname.bib} 

\appendix

\section{Implementation Details} \label{app:implement}
We provide the complete procedures of our framework in Algorithm~\ref{alg:forward} and Algorithm~\ref{alg:online}, and present detailed implementation and settings in the following sections.

\smallsection{Fixed Settings} For all backbones, we adopted the recommended hyperparameter settings provided by their original authors. For consistency across experiments, the input sequence length ($L$) was fixed to $96$. Regarding DynaME's internal components, the gating network's hidden dimension was set to half of the feature dimension $D$, and the regularization coefficient for the linear regression solver was fixed at $\lambda = 10^{-4}$. The safety mechanism only requires spike detection and is insensitive to parameter variations: the exponential smoothing factor is set to $\alpha = 0.95$, while the sensitivity parameter is fixed to $\delta = 0.01$.

\smallsection{Hyperparameter Search Space} The following hyperparameters were subject to grid search: the maximum number of historical samples per expert $n \in \{4, 8, 12, 16\}$, the number of experts $k \in \{1, 2, 3, 4\}$, the minimum fallback $\beta \in \{0.1, 0.2, 0.3\}$, and the history buffer length $M \in \{336, 672\}$.

\smallsection{Codebase References}
The implementation is based on the official public repositories for the following components:

\begin{itemize}[leftmargin=2em]
\item PROCEED \cite{proceed}: \url{https://github.com/SJTU-DMTai/OnlineTSF}
\item xPatch \cite{xpatch}: \url{https://github.com/stitsyuk/xPatch}
\item DSOF \cite{dsof}: \url{https://github.com/yyalau/iclr2025_dsof}
\end{itemize}

\begin{algorithm}[t]
\caption{DynaME: Forward Pass}
\label{alg:forward}
\begin{algorithmic}[1]
\Statex \textbf{Input:} Input $\mathbf{x}$, history $\mathbf{h}$, backbone $\phi$, expert $f_0$, gate $g$.
\Statex \textbf{Output:} Expert predictions $\{\hat{Y}_0, \dots, \hat{Y}_k\}$, weight $\tilde{\omega}$.
\Statex
\Procedure{\textsc{Forward}}{$x, h, \phi, f_0, g$}
    \State $\mathcal{P} \leftarrow \text{IdentifyPeriods}(\mathbf{h})$ 
    \State $\{(X_i, Y_i)\}_{i=1}^k \leftarrow \text{ConstructBatches}(\mathbf{h}, \mathcal{P})$
    \State $\mathbf{z} \leftarrow \phi(\mathbf{x})$
    \State $\hat{\mathbf{y}}_0 \leftarrow f_0(\mathbf{z})$
    \For{$i = 1, \dots, k$}
        \State $Z_i \leftarrow \phi(X_i)$
        \State $\hat{\mathbf{y}}_i \leftarrow (\mathbf{z} Z_i^T) (Z_i Z_i^T + \lambda I)^{-1} Y_i$
    \EndFor
    \State $\tilde{\omega} \leftarrow g(z)$ 
    \State \Return $\{\hat{\mathbf{y}}_0, \dots, \hat{\mathbf{y}}_k\}, \tilde{\omega}$
\EndProcedure
\end{algorithmic}
\end{algorithm} \begin{algorithm}[H]
\caption{DynaME: Online Adaptation and Prediction}
\label{alg:online}
\begin{algorithmic}[1]
\Statex \textbf{Input:} Online data stream, pre-trained backbone $\phi$, general expert $f_0$ and gate $g$.
\State Initialize $\mu^{\text{MSE}}, d$.
\For{time step $t = T_{\text{start}}, \dots $}
    \Statex \hfil \texttt{\textbackslash* 1. Adaptation Step *\textbackslash} \hfil
    \State $\{\hat{\mathbf{y}}_{t-H, i}\}_{i=0}^k, \tilde{\omega} \leftarrow \textsc{Forward}(\mathbf{x}_{t-H}, \mathbf{h}_{t-H}, \phi, f_0, g)$

    \State $\gamma \leftarrow \beta + d_{t-1} (1 - \beta)$
    \State $\omega \leftarrow (1 - \gamma) \tilde{\omega} + \gamma [1, 0, \dots, 0]$
    \State $\hat{\mathbf{y}}_{t-H} \leftarrow \sum_{i=0}^k \omega_{i} \cdot \hat{\mathbf{y}}_{t-H, i}$
    \State $\text{MSE}_t \leftarrow \text{MSE}(\hat{\mathbf{y}}_{t-H}, \mathbf{y}_{t-H})$
    \State Update parameters of $\phi$, $f_0$ and $g$ using $\nabla \text{MSE}_t $.
    \State $\mu_t^{\text{MSE}} \leftarrow (1 - \alpha) \cdot \text{MSE}_t  + \alpha \cdot \mu_{t-1}^{\text{MSE}}$
     \State $d_t \leftarrow 1 - \exp(-\delta (\text{MSE}_t  - \mu_t^{\text{MSE}})^2)$
     
    \Statex
    \Statex \hfil \texttt{\textbackslash* 2. Prediction Step *\textbackslash} \hfil
    \State $\{\hat{\mathbf{y}}_{t, i}\}_{i=0}^k, \tilde{\omega} \leftarrow \textsc{Forward}(x_t, h_t, \phi, g)$
    \State $\gamma \leftarrow \beta + d_t (1 - \beta)$
    \State $\omega \leftarrow (1 - \gamma) \tilde{\omega} + \gamma [1, 0, \dots, 0]$
    \State $\hat{\mathbf{y}}_t \leftarrow \sum_{i=0}^k \omega_{t, i} \cdot \hat{\mathbf{y}}_{t, i}$
    \State \textbf{Output} $\hat{y}_t$
    \EndFor
\end{algorithmic}
\end{algorithm}

\newpage

\section{Dataset Details} \label{app:dataset}
We conducted experiments on several widely used public Time Series Forecasting benchmarks. Table~\ref{tab:stat} summarizes the key statistics, and a brief overview of each dataset is provided below:

\begin{itemize}[leftmargin=2em]
\item \textbf{ETT}\footnote{https://github.com/zhouhaoyi/ETDataset}
 contains oil temperature and six related power load variables, recorded at both hourly and 15-minute resolutions from electricity transformers between 2016 and 2018.
\item \textbf{Electricity}\footnote{\url{https://archive.ics.uci.edu/dataset/321/electricityloaddiagrams20112014}}
 reports hourly electricity consumption for 321 clients over the period from 2012 to 2014.
\item \textbf{Traffic}\footnote{\url{https://pems.dot.ca.gov/}}
 includes hourly occupancy rates collected from loop detectors installed on the San Francisco Bay Area freeways during 2015–2016.
 \item \textbf{Weather}\footnote{\url{https://github.com/zhouhaoyi/Informer2020}}
 comprises 21 meteorological variables such as temperature, humidity, and wind speed, recorded every 10 minutes during 2010-2013.
\end{itemize}

\begin{table}[t]
   \caption{Statistics of the datasets.}\label{tab:stat}
    \renewcommand{\arraystretch}{0.9}
   \centering

    \begin{tabular}{r|ccccc}
    \toprule
    Dataset &  ETTh2 & ETTm1 & Traffic & ECL & Weather \\
    \midrule
    Length    & 17,420 & 69,680 & 17,544  & 26,304  & 52,696  \\
    Channels  & 7     & 7     & 862   & 321 & 21  \\ 
    \bottomrule
    \end{tabular}
   \renewcommand{\arraystretch}{1.0}
\end{table}

\end{document}